\documentclass[runningheads]{llncs}
\usepackage{graphicx}
\usepackage{multirow}
\newcommand{\bhline}[1]{\noalign{\hrule height #1}}
\begin{document}
\title{Biomedical Image Segmentation by Retina-like Sequential Attention Mechanism
Using Only A Few Training Images}
\titlerunning{Biomedical Image Segmentation by Retina-like Sequential Attention}
%
\author{Shohei Hayashi
\and Bisser Raytchev\thanks{corresponding author, email: bisser@hiroshima-u.ac.jp}
\and Toru Tamaki
\and Kazufumi Kaneda}
\institute{Department of Information Engineering, Hiroshima University, Japan}
%
%
%
\maketitle              
\begin{abstract}
In this paper we propose a novel deep learning-based algorithm for biomedical image segmentation
which uses a sequential attention mechanism able to shift the focus of attention across the image in a
selective way, allowing subareas which are more difficult to classify to be processed at increased resolution. 
The spatial distribution of class information in each subarea is learned using a retina-like representation where resolution decreases with distance from the center of attention.  
The final segmentation is achieved by averaging class predictions over overlapping subareas, utilizing the power of ensemble learning to increase segmentation accuracy.
Experimental results for  semantic segmentation task
for which only a few training images are available show that a CNN using the proposed method outperforms both a
patch-based classification CNN and a fully convolutional-based method.
\end{abstract}
\section{Introduction}
Recently deep learning methods \cite{DLbook}, which automatically extract hierarchical features 
capturing complex nonlinear relationships in the data, have managed to  
successfully replace most task-specific hand-crafted features. This has resulted in a significant
improvement in performance on a variety of biomedical image analysis tasks, like 
object detection, recognition and segmentation
(see e.g. \cite{DLMIAbook} for a recent survey of the field and representative methods used in different applications),
and currently Convolutional Neural Network (CNN) based methods define the state-of-the-art in this area.

In this paper we concentrate on biomedical image \emph{segmentation}. For segmentation, where each pixel needs 
to be classified into its corresponding class, initially \emph{patch-wise training/classification} was used \cite{patch}. 
In patch-based methods, local patches of pre-determined size are extracted 
from the images, typically using a CNN as pixel-wise classifier. During training, the patch is used as an input to the network and it is assigned as a label the class of the pixel at the \emph{center} of the patch (available from ground-truth
data provided by a human expert). During the test phase, a patch is fed into the trained net and the output layer of the
net provides the probabilities for each class. More recently, Fully Convolutional Networks (FCN) \cite{FCN}, which 
replace the fully connected layers with convolutional ones, have
replaced the patch-wise approach by providing a more efficient way to train CNNs end-to-end, pixels-to-pixels, and methods stemming from this approach presently define the state-of-the-art in biomedical image segmentation, exemplified by conv-deconv-based methods like U-Net \cite{unet}.

\begin{figure}[ht]
  \centering
  \includegraphics[width=\linewidth]{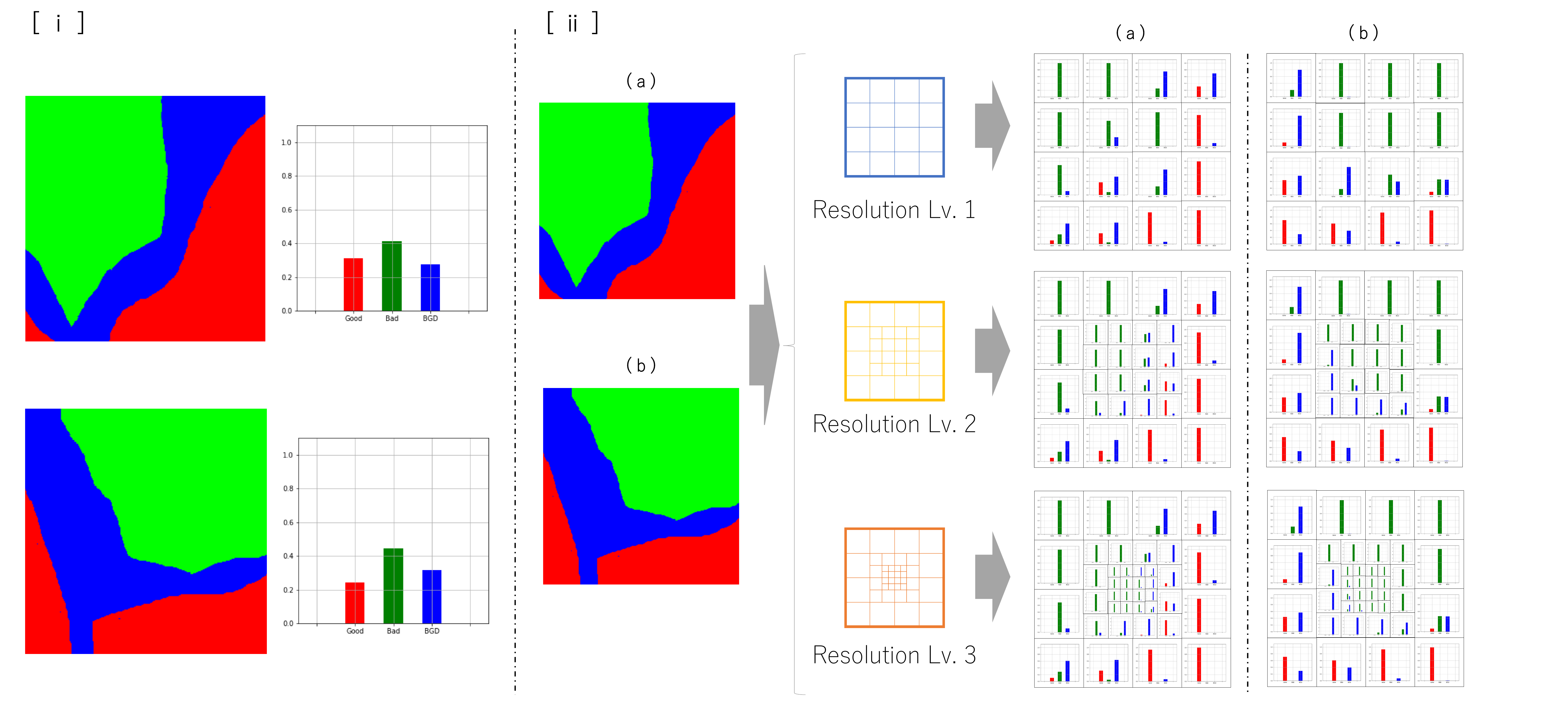}
  \caption
  {[i] Local patch containing pixels belonging to 3 classes shown in different color.
  On the right is normalized histogram showing the empirical distribution of the classes inside the patch, which can be interpreted as probabilities.
  A standard patch-wise method learns only the class of the pixel at the center, ignoring completely class topology.
  [ii] The proposed method learns the structure of the spatial distribution of the
  class topology in a local subarea represented similarly to the retina - the
  resolution is highest in the center and decreases progressively in the periphery.
  As attention shifts inside the image, the information of overlapping subareas is
  combined to produce the final segmented image (details explained in the text).
  Figure best viewed in color.}
  \label{res_patch}
\end{figure}

  Although fully convolutional methods have shown state-of-the-art performance on many segmentation tasks, they
typically need to be trained on large datasets to achieve good accuracy. In many biomedical image segmentation tasks, however, only
a few training images are available -- either data simply being not available, or providing pixel-level ground truth 
by experts being too costly to obtain. Here we are motivated by a similar problem (section 3), 
where less than 50 images are available for training.
On the other hand, \emph{patch-wise methods} need only local patches, a huge number of which can be extracted 
even from a small number of training images. They however suffer from the following problem. While fully convolutional methods
learn a map from pixel areas (multiple input image values) to pixel areas (multiple classes of all the pixels in the area),
patch-wise methods learn a map from pixel areas (input image values) to a single pixel (class of the pixel in the center  of the patch). 
As illustrated in Fig.~\ref{res_patch} [i], 
this wastes the rich information about the topology 
of the class structure inside the patch for many of the samples which contain more than a 
single class, and these would typically be the most interesting/difficult samples \cite{topology}. 
Instead of trying to represent in a suitable way and learn the complex class topology, it just substitutes it by
a single class (the class of the pixel in the center of the patch). 

Regarding fully convolutional methods, they treat all locations in the images 
in the same way, which is in contrast with how human visual perception works.
It is known that humans employ attentional mechanisms to focus selectively on
\emph{subareas of interest} and construct a global scene-representation by combining the information from different local subareas \cite{scenes}.  

Based on the above observations, we propose a new method, which takes a middle ground between fully convolutional and patch-wise learning and combines the benefits of both 
of these strategies.
As shown in Fig.~\ref{res_patch}, as in the patch-wise approach we consider subareas
of the whole image at a time, which provides us with sufficient number of training
samples, even if only a few ground-truth labeled images are available. 
However, as illustrated in Fig.~\ref{res_patch} [ii], the class information is organized
as in the retina \cite{kandel}: the spatial resolution is highest in the central area (corresponding to the \emph{fovea}), and it
diminishes as we go to the periphery of the subarea.
We propose a \emph{sequential attention mechanism} which shifts the focus of attention
in such a way that areas of the image which are difficult to classify (i.e. the
classification uncertainty is higher) are considered in much more detail than
areas which are easy to classify. Since the focus of attention moves the subarea 
under investigation much slower over difficult areas (i.e. with much smaller step), this results in many \emph{overlapping
subareas} in these regions. The final segmentation is achieved by averaging the
class predictions over the overlapping subareas. In this way, the power of \emph{ensemble
learning}  \cite{ensembles} is utilized by incorporating information from all 
overlapping subareas in the neighborhood (each of which provides slightly different 
views of the scene) to further improve accuracy. 
  
This is the basic
idea of the method proposed in the paper, and details how to implement it in a CNN will be given in the next section. 
Experimental results are reported in section 3, indicating that a significant improvement 
in segmentation accuracy can be achieved by the proposed method compared to both patch-based and fully convolutional-based methods. 

\section{Method}
  We represent a subarea $\cal{S}$ extracted from an input image (centered at
  the current focus of attention) as a tensor of size $d \times d \times c$,
  where $d$ is the size of the subarea in pixels and $ c $ stands for the color channels if color images are used (as typically done, and $ c = 3 $ for RGB images).
  As shown in Fig. \ref{res_patch} [ii], we can represent the class information
  corresponding to this subarea as grids of different resolution, where each
  cell in a grid contains a 
sample histogram $h^{(i)}$ calculated so that the $k$-th element of  $h^{(i)}$ 
gives the number of pixels from class $k$ observed in the $i$-th cell of the grid.
If each histogram $h^{(i)}$ is normalized to sum to 1 by dividing each bin by the total number of 
pixels covered by the cell, the corresponding vector  $p^{(i)}$ can now be used to represent the 
probability mass function (\emph{pmf}) for cell $i$: the $k$-th element of  $p^{(i)}$, i.e. 
$p_k^{(i)}$, 
can be interpreted as the probability of observing class $k$ in the area covered by
the $i$-th cell of the grid. 
  
Next, we show how retina-like grids of different resolution levels can be created.
Let's start with a grid of size $4 \times 4$, as shown in the top row of Fig. \ref{res_patch} [ii].
This we will call \emph{Resolution Level} 1 and denote it as $r = 1$. At resolution level 1 all
the cells in the grid have the same resolution, i.e. the \emph{pmf}s corresponding to each
cell are calculated from areas of the same size. For example, if the size of the local subarea
under consideration is $d = 128$ pixels (i.e. image patch of size $128 \times 128$ pixels), each cell in the grid at
resolution level $r = 1$ would correspond to an image area of size $32 \times 32$ pixels from which
a probability mass function $p^{(i)}$ would be calculated, as explained above.
Next, we can create a grid at \emph{Resolution Level} 2 ($r = 2$) by dividing in half 
the four cells in the center of the grid, so that they form an inner $4 \times 4$ grid, whose
resolution is double. We can continue this process of dividing the innermost 4 cells into 2
to obtain still higher resolution levels. It is easy to see that the number of cells $N$ in a grid
obtained at resolution level $r$ is $N = 16 + 12(r-1)$. 
Of course, it is not necessary the initial grid at $r = 1$ to be of 
size $4 \times 4$, but choosing this number makes the process of creating different resolution
levels especially simple, since in this case the innermost cells are always 4 ($2 \times 2$).

In our method, we train a CNN to learn the map between a local subarea image given as input to the network, and the corresponding \emph{pmf}s $p^{(i)}$ used as target values.
We use the cross-entropy between the \emph{pmf}s of the targets $p^{(i)}$ and the
corresponding output unit activations $y^{(i)}$ as loss function $L$:
  \begin{equation}
    L = - \sum_{n}\sum_{i}\sum_{k} p^{(i)}_{k,n} \log{y^{(i)}_{k,n}} (S_n ; \vec{w}),
    \label{equ:cost}
  \end{equation}
  where $n$ indexes the training subarea image patches ($S_n$ being the $n$-th training subarea image
patch), $i$ indexing the cells in the corresponding resolution grid, and $k$ the classes.
Here, $\vec{w}$ represents the weights of the network, to be found by minimizing the loss function. To form probabilities, the network output units corresponding to each
cell are passed through the \emph{soft-max} activation function.
  
\begin{figure}[t]
  \centering
  \includegraphics[width=\linewidth]{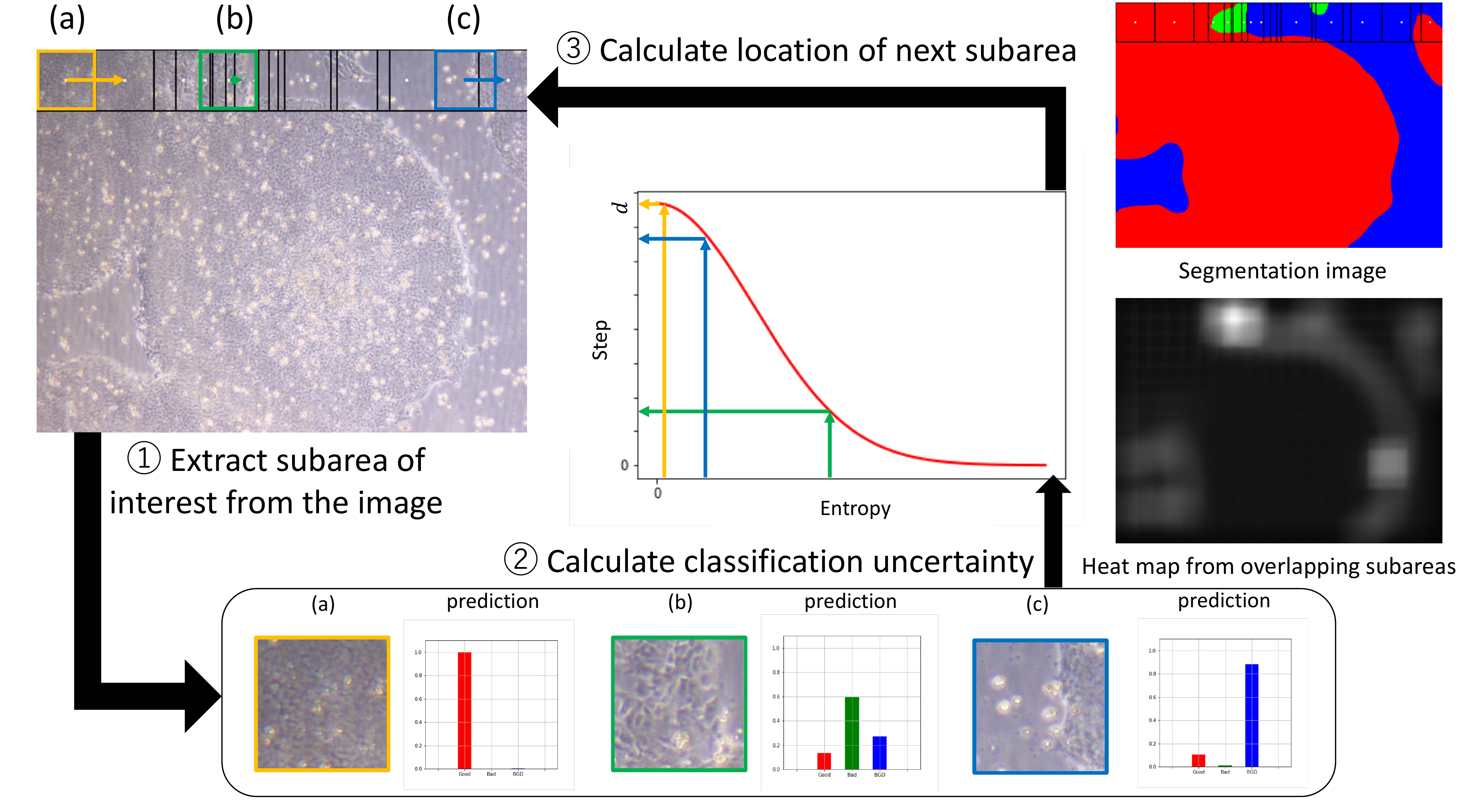}
  \caption
  {Overview of the sequential attention mechanism (see text for details).
}
  \label{attention_step}
\end{figure}

  Finally, we describe the sequential attention mechanism we utilize, whose purpose is to move
the focus of attention across the image in such a way that those parts which are difficult
to classify (i.e. classification uncertainty is high) are observed at the highest possible
resolution, and the retina-like grid of \emph{pmf}s moves with smaller steps across such areas.
To evaluate the classification uncertainty of the grid over the present subarea $\cal{S}$,
we use the following function,  
  \begin{equation}
    H(S) = - \frac{1}{N}\sum_{i \in S}\sum_{k} p^{(i)}_{k} \log{p^{(i)}_{k}},
    \label{equ:entropy}
  \end{equation}
  which represents the average entropy obtained from the 
posterior \emph{pmf}s $p^{(i)}$ for each cell (indexed by $i$) inside the grid, and
$k$ indexes the classes. Using $H(S)$ as a measure of classification uncertainty, the position
of the next location where to move the \emph{focus of attention} (horizontal shift in pixels) is given by
  \begin{equation}
    f(H(S)) = d \exp\{{-(H(S))^2/2\sigma^2}\}.
    \label{equ:gauss}
  \end{equation}
The whole process is illustrated in Fig.~\ref{attention_step}.
We start at the upper left corner of the input image, with a subarea of size $d \times d$ pixels
(the yellow patch (a) in the figure). The classification uncertainty for that subarea is calculated
using Eq.~\ref{equ:entropy}, and the step size in pixels to move in the horizontal direction
is calculated by Eq.~\ref{equ:gauss}. As illustrated in the graph in the center of Fig.~\ref{attention_step}, since in this case the classification uncertainty is 0
(all pixels in the subarea belong to the same class), the focus of attention moves $d$ pixels
to the right, i.e. in this extreme case there is no overlap between the current and next subareas.
For the subarea (b) shown in green, the situation is very different. In this case the 
classification uncertainty is very high, and the focus of attention would move only slightly
to the right, allowing the image around this area to be assessed at the highest resolution.
This would result in very high level of overlap between neighboring subareas, as shown in the
heat map on the right (where intensity is proportional to the level of overlap). This process is repeated until the right corner of the image is reached.
Then the focus of attention is moved 10 pixels in the vertical direction to scan
the next row and everything is repeated until
the whole image is processed.

  While the above attention mechanism moves the focus of attention across the image, 
the posterior class \emph{pmf}s from the grids corresponding to each subarea
are stored in a \emph{probability map} of the same size as the image, i.e. to each pixel in
the image is allocated a \emph{pmf} equal to the \emph{pmf} of the cell from the grid positioned
above that pixel. In areas in the image where several subareas overlap, the probability map 
is computed by averaging for each pixel the \emph{pmf}s of all cells which partially overlap over that pixel. 
Finally, the class of the pixel is
obtained by taking the class with highest probability from the final probability map, as
shown in the upper right corner of Fig.~\ref{attention_step} for the final segmented image.
  
      \begin{table}[t]
       \centering
        \caption{Experimental results for different values $d$ of the size of the local subareas }
        \label{score}
        \scalebox{0.85}{
        \begin{tabular}{c|c|c|c|c|c|c}
            $d$ & Method  & Jaccard Index  & Dice  &  TPR  &  TNR  &  Accuracy \\
            \hline
            \hline
             \multirow{6}{*}{96}  &  patch-center &  0.811 $\pm$ 0.039  &  0.879 $\pm$ 0.034  &  0.877 $\pm$ 0.039  &  0.922 $\pm$ 0.014  &  0.932 $\pm$ 0.016 \\  \cline{2-7}
            &ResLv-1 &  0.798 $\pm$ 0.046  &  0.868 $\pm$ 0.040  &  0.866 $\pm$ 0.042  &  0.914 $\pm$ 0.016  &  0.928 $\pm$ 0.015 \\  \cline{2-7}
            &ResLv-2 &  0.811 $\pm$ 0.037  &  0.879 $\pm$ 0.031  &  0.878 $\pm$ 0.035  &  0.919 $\pm$ 0.020  &  0.931 $\pm$ 0.015 \\ \cline{2-7}
            &ResLv-3 &  0.819 $\pm$ 0.034  &  0.884 $\pm$ 0.029  &  0.883 $\pm$ 0.035  &  0.923 $\pm$ 0.012  &  0.936 $\pm$ 0.010 \\ \cline{2-7}
            &ResLv-4 &  0.820 $\pm$ 0.033  &  0.884 $\pm$ 0.029  &  0.881 $\pm$ 0.036  &  0.926 $\pm$ 0.013  &  0.936 $\pm$ 0.012 \\ \cline{2-7}
            &UNet-patch &  0.788 $\pm$ 0.051  &  0.861 $\pm$ 0.047  &  0.859 $\pm$ 0.045  &  0.916 $\pm$ 0.013  &  0.922 $\pm$ 0.011 \\ \cline{2-7}
            \bhline{1pt}
             \multirow{7}{*}{128}  & patch-center &  0.811 $\pm$ 0.029  &  0.878 $\pm$ 0.026  &  0.873 $\pm$ 0.028  &  0.917 $\pm$ 0.012  &  0.931 $\pm$ 0.012 \\  \cline{2-7}
            &ResLv-1 &  0.812 $\pm$ 0.038  &  0.878 $\pm$ 0.032  &  0.874 $\pm$ 0.033  &  0.918 $\pm$ 0.020  &  0.935 $\pm$ 0.012 \\  \cline{2-7}
            &ResLv-2 &  0.815 $\pm$ 0.039  &  0.881 $\pm$ 0.032  &  0.878 $\pm$ 0.034  &  0.923 $\pm$ 0.015  &  0.934 $\pm$ 0.014 \\ \cline{2-7}
            &ResLv-3 &  0.816 $\pm$ 0.030  &  0.881 $\pm$ 0.029  &  0.879 $\pm$ 0.033  &  0.920 $\pm$ 0.011  &  0.935 $\pm$ 0.008 \\ \cline{2-7}
            &ResLv-4 &  0.818 $\pm$ 0.034  &  0.821 $\pm$ 0.031  &  0.881 $\pm$ 0.031  &  0.921 $\pm$ 0.017  &  0.936 $\pm$ 0.011 \\ \cline{2-7}
            &ResLv-5 &  0.826 $\pm$ 0.032  &  0.891 $\pm$ 0.030  &  0.890 $\pm$ 0.032  &  0.924 $\pm$ 0.018  &  0.936 $\pm$ 0.009 \\ \cline{2-7}
            &UNet-patch &  0.810 $\pm$ 0.035  &  0.879 $\pm$ 0.030  &  0.873 $\pm$ 0.034  &  0.921 $\pm$ 0.014  &  0.933 $\pm$ 0.012 \\ \cline{2-7}
            \bhline{1pt}
             \multirow{7}{*}{192}  &  patch-center &  0.778 $\pm$ 0.029  &  0.856 $\pm$ 0.025  &  0.849 $\pm$ 0.020  &  0.899 $\pm$ 0.017  &  0.917 $\pm$ 0.017 \\  \cline{2-7}
            &ResLv-1 &  0.810 $\pm$ 0.037  &  0.876 $\pm$ 0.030  &  0.872 $\pm$ 0.031  &  0.914 $\pm$ 0.021  &  0.933 $\pm$ 0.016 \\  \cline{2-7}
            &ResLv-2 &  0.817 $\pm$ 0.041  &  0.880 $\pm$ 0.038  &  0.879 $\pm$ 0.041  &  0.922 $\pm$ 0.017  &  0.936 $\pm$ 0.011 \\ \cline{2-7}
            &ResLv-3 &  0.821 $\pm$ 0.032  &  0.885 $\pm$ 0.030  &  0.883 $\pm$ 0.032  &  0.926 $\pm$ 0.010  &  0.935 $\pm$ 0.011 \\ \cline{2-7}
            &ResLv-4 &  \bf{0.832 $\pm$ 0.036}  &  \bf{0.894 $\pm$ 0.030}  &  \bf{0.890 $\pm$ 0.034}  &  \bf{0.926 $\pm$ 0.015}  &  \bf{0.940 $\pm$ 0.012} \\ \cline{2-7}
            &ResLv-5 &  0.825 $\pm$ 0.032  &  0.887 $\pm$ 0.030  &  0.883 $\pm$ 0.032  &  0.925 $\pm$ 0.015  &  0.938 $\pm$ 0.012 \\ \cline{2-7}
            &UNet-patch &  0.809 $\pm$ 0.036  &  0.878 $\pm$ 0.029  &  0.870 $\pm$ 0.034  &  0.920 $\pm$ 0.015  &  0.933 $\pm$ 0.015 \\ \cline{2-7}
            \bhline{1pt}
             \multirow{7}{*}{256}  &  patch-center &  0.732 $\pm$ 0.038  &  0.822 $\pm$ 0.037  &  0.813 $\pm$ 0.039  &  0.862 $\pm$ 0.023  &  0.897 $\pm$ 0.019 \\  \cline{2-7}
            &ResLv-1 &  0.804 $\pm$ 0.039  &  0.871 $\pm$ 0.035  &  0.866 $\pm$ 0.036  &  0.910 $\pm$ 0.018  &  0.931 $\pm$ 0.012 \\  \cline{2-7}
            &ResLv-2 &  0.810 $\pm$ 0.038  &  0.877 $\pm$ 0.037  &  0.870 $\pm$ 0.040  &  0.918 $\pm$ 0.018  &  0.932 $\pm$ 0.012 \\ \cline{2-7}
            &ResLv-3 &  0.819 $\pm$ 0.029  &  0.882 $\pm$ 0.028  &  0.879 $\pm$ 0.034  &  0.922 $\pm$ 0.016  &  0.937 $\pm$ 0.010 \\ \cline{2-7}
            &ResLv-4 &  0.814 $\pm$ 0.033  &  0.877 $\pm$ 0.030  &  0.871 $\pm$ 0.031  &  0.917 $\pm$ 0.020  &  0.936 $\pm$ 0.011 \\ \cline{2-7}
            &ResLv-5 &  0.815 $\pm$ 0.038  &  0.880 $\pm$ 0.034  &  0.876 $\pm$ 0.035  &  0.919 $\pm$ 0.016  &  0.934 $\pm$ 0.012 \\ \cline{2-7}
            &UNet-patch &  0.811 $\pm$ 0.034  &  0.879 $\pm$ 0.028  &  0.874 $\pm$ 0.030  &  0.921 $\pm$ 0.014  &  0.933 $\pm$ 0.012 \\ 
            \bhline{1pt}
            & UNet-image &  0.806 $\pm$ 0.033  &  0.877 $\pm$ 0.029  &  0.874 $\pm$ 0.031  &  0.919 $\pm$ 0.013  &  0.928 $\pm$ 0.008 \\ 
        \end{tabular}
        }
    \end{table}

  \begin{center}
  \begin{figure}[t]
  \includegraphics[width=\linewidth]{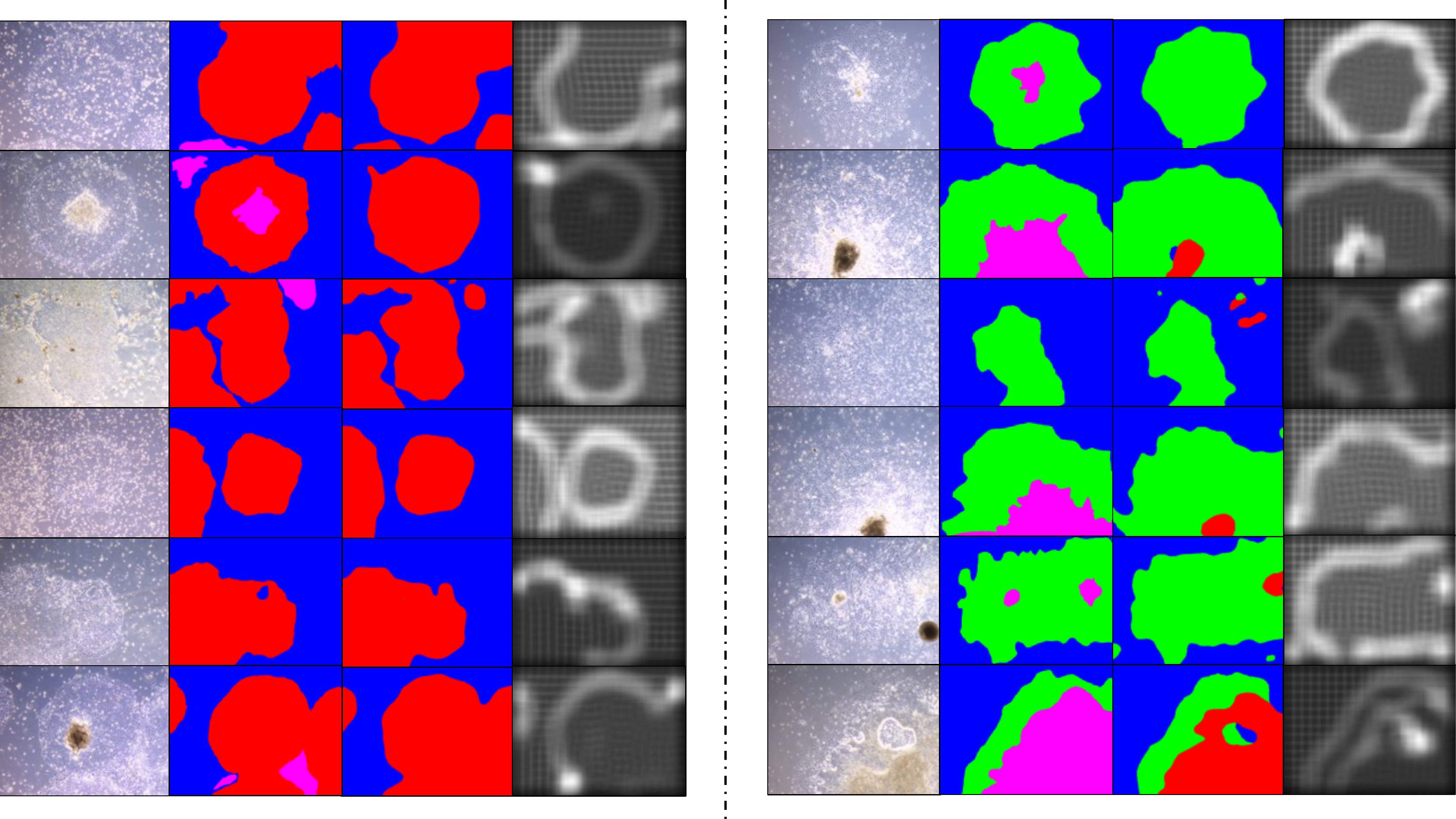}
  \caption
  {Segmentation results for several images from the iPS dataset, obtained by the proposed method (3rd column), using ResLv-4 with subarea size of $d = 192$.
  First column shows the original images and second column the ground truth segmentation provided by
  an expert (red corresponds to class Good, green to Bad and blue to Background). The last column shows the corresponding heat map, where areas in which there was
  high overlap over neighboring subareas are shown with high intensity values.
  }
  \label{result_image}
  \end{figure}
  \end{center}

  \section{Experiments}
  In this section we evaluate the proposed method in comparison with a standard patch-wise 
  classification-based CNN \cite{patch} and the fully convolutional-based U-Net \cite{unet} on the dataset described below.
  Additionally we implemented a U-Net version, called \emph{UNet-patch}, which applies U-Net to local patches rather than to a whole image. The original U-Net method which takes as input the whole image we will call \emph{UNet-image}. \\
  {\bf Dataset: } Our dataset consists of 59 images showing colonies of undifferentiated and
differentiated iPS cells obtained through phase-contrast microscopy. 
Induced pluripotent stem (iPS) cells \cite{ips}, for whose discovery S. Yamanaka received the 
Nobel prize in Physiology and Medicine in 2012, contain great promise for regenerative medicine.
Still, in order to fulfill their promise a
steady supply of iPS cells obtained through harvesting of individual cell colonies
is needed and automating the detection of abnormalities arising during the cultivation process is crucial.
Thus, our task is to segment the input images into three categories: Good (undifferentiated), Bad (differentiated) and Background (BGD, the culture medium).
  Several representative images together with ground-truth provided by experts can be seen in 
  Fig.~\ref{result_image}.
  All images in this dataset are of size $1600 \times 1200$ pixels. Several images contained a few locations where even the experts were not sure what the corresponding class was.
  Such ambiguous regions are shown in pink and these areas were not used during training and not evaluated during test. \\
{\bf Network Architecture and Hyperparameters:} We used a network architecture based on the VGG-16 CNN net, apart from the following differences.
  There are 13 convolutional layers in VGG-16, while we used 10 here. Also, in VGG-16 there are 3 fully-convolutional layers of which the first two consist of 4096 units,
  while those had 1024 units in our implementation.
  The learning rate was set to 0.0001 and for the optimization procedure we used ADAM.
  Batch size was 16, training for 20 epochs (U-Net-patch for 15 epochs and U-Net-image for 200 epochs). For the implementation of the CNNs we used TensorFlow and Keras.
  Four different sizes for the local subareas were tried: $d = 96, 128, 192, 256$ and resolution level was changed between $r = 1$ to $r = 5$. The width of the Gaussian in Eq.~\ref{equ:gauss} was empirically set to $\sigma = 0.4$ for all experimental results.
{\bf Evaluation procedure and criteria:} The quality of the obtained segmentation results for each method were evaluated by 5-fold cross-validation using the following criteria: 
  Jaccard index (the most challenging one), Dice coefficient, True Positive Rate (TPR), True Negative Rate (TNR) and Accuracy. For each score the average and standard deviation are reported. \\
{\bf Results:} The results obtained on the iPS cell colonies dataset for each of the methods
are given in Table \ref{score},
where, \emph{patch-center} stands for the patch-wise classification method, and
results for resolution levels from $r = 1$ to $r = 5$ are given for the proposed method.
As can be seen from the results, the proposed method outperforms both patch-wise classification
and the U-Net-based methods.
  Fig.~\ref{result_image} gives some examples of segmentation on images from the iPS
dataset, showing that very good accuracy of segmentation can be achieved by the proposed method. 
The heat maps given in the last column demonstrate that the proposed attentional mechanism
is able to focus the high-resolution parts of the retina-like grid on the boundaries between 
the classes which seem to be most difficult to classify, resulting in increased accuracy
of segmentation.

%

  \section{Conclusion}
  In this paper we have shown that the combined power of (1) a sequential attention mechanism controlling the shift of the focus of attention, (2) local retina-like representation of the spatial 
  distribution of class information and (3) ensemble learning can lead to increased segmentation accuracy in biomedical segmentation tasks. We expect that the proposed method can be especially useful for datasets for which only a few training images are available.
%
%
%
%
\bibliographystyle{splncs04}
\bibliography{miccai2019}

\begin{thebibliography}{10}
\providecommand{\url}[1]{\texttt{#1}}
\providecommand{\urlprefix}{URL }
\providecommand{\doi}[1]{https://doi.org/#1}

\bibitem{patch}
Cire\c{s}an, D.C., Giusti, A., Gambardella, L.M., Schmidhuber, J.: Deep neural
  networks segment neuronal membranes in electron microscopy images. In: NIPS.
  pp. 2843--2851 (2012)

\bibitem{DLbook}
Goodfellow, I., Bengio, Y., Courville, A.: Deep Learning. MIT Press (2016)

\bibitem{kandel}
Kandel, E.R., Schwarz, J., Jessell, T.M., Siegelbaum, S.A., Hudspeth, A.J.:
  Principles of Neural Science, 5ed. McGraw-Hill (2013)

\bibitem{topology}
Kontschieder, P., Bulo, S.R., Bischof, H., Pelillo, M.: Structured class-labels
  in random forests for semantic image labeling. In: Proc. ICCV2012. pp. 2190
  -- 2197 (2012)

\bibitem{ensembles}
Kuncheva, L.: Combining Pattern Classifiers, 2ed. Wiley (2014)

\bibitem{scenes}
Rensink, R.A.: The dynamic representation of scenes. Visual Cognition
  \textbf{7}(1-3),  17--42 (2000)

\bibitem{unet}
Ronneberger, O., Fischer, P., Brox, T.: U-net: Convolutional networks for
  biomedical image segmentation. In: MICCAI (2015)

\bibitem{FCN}
Shelhamer, E., Long, J., Darrell, T.: Fully convolutional networks for semantic
  segmentation. {IEEE} Trans. Pattern Anal. Mach. Intell.  \textbf{39}(4),
  640--651 (2017)

\bibitem{ips}
Takahashi, K., Tanabe, K., Ohnuki, M., Narita, M., Ichisaka, T., Tomoda, K.,
  Yamanaka, S.: Induction of pluripotent stem cells from adult human
  fibroblasts by defined factors. Cell  \textbf{131 (5)},  861--871 (2007)

\bibitem{DLMIAbook}
Zhou, S.K., Greenspan, H., Shen, D. (eds.): Deep Learning for Medical Image
  Analysis. Academic Press (2017)

\end{thebibliography}
%
\end{document}